\documentclass[a4paper,conference]{IEEEtran}
\usepackage{times}
\usepackage{soul}
\usepackage{url}
\usepackage[hidelinks]{hyperref}
\usepackage[utf8]{inputenc}
\usepackage[small]{caption}
\usepackage{graphicx}
\usepackage{amsmath}
\usepackage{booktabs}
\usepackage{algorithm}
\usepackage{algorithmic}
\usepackage{multirow}

\begin{document}

\title{Adaptive Remote Sensing Image Attribute Learning \\for Active Object Detection}


\author{\IEEEauthorblockN{Nuo Xu\IEEEauthorrefmark{1},
Chunlei Huo\IEEEauthorrefmark{1},
Jiacheng Guo\IEEEauthorrefmark{2},
Yiwei Liu\IEEEauthorrefmark{3},
Jian Wang\IEEEauthorrefmark{4} and
Chunhong Pan\IEEEauthorrefmark{1}}
\IEEEauthorblockA{\IEEEauthorrefmark{1}NLPR, Institute of Automation, Chinese Academy of Sciences, \\
School of Artificial Intelligence, University of Chinese Academy of Sciences, Beijing, China \\ Email: nuo.xu@nlpr.ia.ac.cn, clhuo@nlpr.ia.ac.cn, chpan@nlpr.ia.ac.cn}
\IEEEauthorblockA{\IEEEauthorrefmark{2}Beijing Information Science and Technology University, Beijing, China. Email: 1941537783@qq.com}
\IEEEauthorblockA{\IEEEauthorrefmark{3}Beijing University of Civil Engineering and Architecture, Beijing, China. Email: 1603375427@qq.com}
\IEEEauthorblockA{\IEEEauthorrefmark{4}College of Robotics, Beijing Union University, Beijing, China. Email: 1552543455@qq.com}}

\maketitle

\begin{abstract}
  In recent years, deep learning methods bring incredible progress to the field of object detection. However, in the field of remote sensing image processing, existing methods neglect the relationship between imaging configuration and detection performance, and do not take into account the importance of detection performance feedback for improving image quality. Therefore, detection performance is limited by the passive nature of the conventional object detection framework. In order to solve the above limitations, this paper takes adaptive brightness adjustment and scale adjustment as examples, and proposes an active object detection method based on deep reinforcement learning. The goal of adaptive image attribute learning is to maximize the detection performance. With the help of active object detection and image attribute adjustment strategies, low-quality images can be converted into high-quality images, and the overall performance is improved without retraining the detector.
\end{abstract}

\IEEEpeerreviewmaketitle

\section{Introduction}
Object detection is one of the fundamental and important issues in the field of computer vision. Remote sensing image object detection is a process of detecting the location of the object of interest from a remote sensing image and identifying its category. Because of the difficulty of feature extraction, position regression, and object classification, it is very challenging. Deep learning technology is a powerful method that can automatically learn feature representation from data, which has developed very rapidly in recent years. The method of deep learning has demonstrated extraordinary effects in solving difficulties of object detection, prompting many efficient methods and outperforming traditional methods. The deep learning method parses the input data by constructing a hierarchical nonlinear learning unit to learn the end-to-end mapping between the image and its semantic label. It is this learning method that has made a great breakthrough in the field of remote sensing image object detection.

Although deep learning is very effective, traditional object detection methods are limited due to the passive nature. Firstly, in the process of in-orbit imaging, the image acquisition process is based on the human visual perception as a reference for checking its quality, and it does not consider the specific requirements of tasks such as object detection. Secondly, the images are directly used for training or testing without proper image quality evaluation, or the images are simply evaluated and manually pre-processed by the visual inspection. In short, evaluating the acquired image in terms of human visual perception is not necessarily the optimal for object detection task. In fact, there is a gap in imaging configuration requirements for visual inspection and object detection, and such difference will impact the performance of the detection model. In other words, adaptive image attribute learning is very important for the in-orbit imaging procedure or the subsequent object detection step, however, it is rarely considered in the literature. Image attributes mean spatial resolution, color, scale (the ratio of the distance on an image to the distance on the ground), hue, saturation, brightness and so on. This paper only takes brightness and scale learning as examples.

\begin{figure}[tb]
  \centering
  \includegraphics[width=8.8cm]{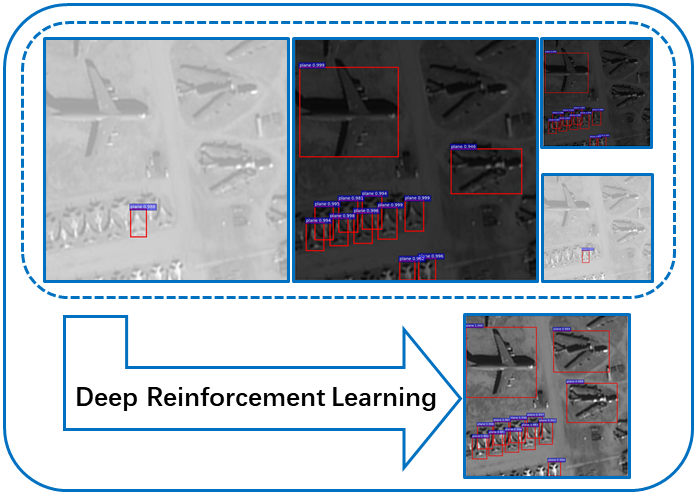}
\caption{Motivation. Due to the limitation of imaging configuration and environmental changes, the detection performance of low-quality images is not good. Therefore, it is necessary to adaptively learn image attributes to improve detection performance.}
\label{fig:fig1}
\end{figure}

In order to overcome the above limitations, this paper proposes an active object detection method based on deep reinforcement learning. The role of reinforcement learning is to optimize imaging conditions and improve object detection performance. It is worth noting that the application of deep reinforcement learning in image processing is a new topic, and the proposed method is different from traditional detection models described in the next section. The novelty of this paper is the combination of deep reinforcement learning with the current mainstream object detection algorithms. By adjusting the image attribute, the image quality is actively improved to adapt to the well-trained detectors, thus, the detection performance will be improved, as shown in Fig. \ref{fig:fig1}. In short, it is useful for offline detection and online imaging.

For convenience, the framework in this paper is named active object detection with reinforcement learning (\emph{RL-AOD}). The most important difference between \emph{RL-AOD} and the mainstream method is that the mainstream detector locates the object in one step through the regression algorithm, but \emph{RL-AOD} can adaptively select the appropriate brightness and scale through sequence decision-making in the process of locating the object. This method can adaptively learn the image attributes with the best object detection performance, which is of great significance for remote sensing image object detection. Our contributions in this paper are summarized as follows:

(1) An active object detection framework \emph{RL-AOD} is proposed, by combining deep reinforcement learning and mainstream deep learning object detection method. This method is used to solve the problem that the imaging configuration and detection tasks do not match.

(2) This paper proposes strategies for adaptively adjusting brightness and scale of images, and combines them together to improve the detection performance of low-quality images.

\section{Related Work}
Active object detection consists of two parts, reinforcement learning and object detection. For convenience, related technologies will be reviewed below.

\textbf{Object Detection.} Before detectors based on deep learning are proposed, DPM \cite{pedro2010object} is a very successful object detection algorithm. DPM method first calculates the gradient direction histogram, then uses SVM (Surpport Vector Machine) to train to obtain the gradient model of the object, and finally uses target matching technology to detect specific objects. Object detection based on deep learning includes position regression and object classification. In the past few years, various new algorithms have constantly been proposed, and there is a strong correlation between these algorithms. In general, deep learning based detection models can be divided into the following two categories, two-stages methods (e.g., Faster RCNN \cite{ren2015faster}, FPN \cite{lin2017feature}, R-FCN \cite{dai2016r}, Cascade RCNN \cite{cai2018cascade}) and one-stage methods (e.g., YOLO \cite{redmon2017yolo9000}, SSD \cite{liu2016ssd}, DSSD \cite{fu2017dssd}, RetinaNet \cite{lin2017focal}, CornerNet \cite{law2018cornernet}). The main difference between the two-stages framework and the one-stage ones is the pre-processing step for generating regional proposals. The former pays more attention to precision, while the latter pays more attention to speed. Compared to two-stage detectors, one-stage detectors simplify the detection process and increasing the speed since regression and classification are performed only once,  and the accuracy is thus being impacted. Future trends will focus more on the combination of the two (e.g., RefineDet \cite{zhang2018single}, RFBNet \cite{liu2018receptive}). This type of method has two or more regression and classification processes. Not only is the accuracy not worse than the two-stage method, but the speed can also be close to the one-stage method. Despite the great success of deep learning, there is still a huge gap between the performance of the current best methods and requirements from practical applications. Traditional object detection methods only passively detect the object, but cannot actively learn the attributes (brightness, scale, etc.) of images. Traditional active object recognition method (e.g., \cite{denzler2002information,wilkes1992active}) is to perform viewpoint control by controlling the camera. The method in this paper does not directly control the camera, but learns images' attributes.

\textbf{Deep Reinforcement Learning.} Reinforcement learning (RL) is a powerful and effective tool for an agent to learn how to make serial sequence decisions based on the external environment. The overall decision made by the agent will be optimal since RL aims to maximize the accumulating rewards. In recent years, traditional RL algorithms have been incorporated into the deep learning framework, thus producing a series of deep reinforcement learning (DRL) models (e.g. DQN \cite{mnih2015human}, DDPG \cite{lillicrap2015continuous}, TRPO \cite{schulman2015trust}), which outperform traditional RL methods. In the early days, RL methods are mainly used for robot control \cite{kormushev2010robot,hester2010generalized}. In recent years, DRL methods have been successfully applied in many fields such as game agents \cite{silver2016mastering,silver2017mastering} and neural network architecture design \cite{baker2017designing,zoph2017neural}. DRL has also attracted people's attention in the field of computer vision (CV). For examples, some scholars use DRL to continuously narrow the detection window to the final object by sequence decision \cite{caicedo2015active,bellver2016hierarchical,jie2016tree}. In detail, they use DRL method alone to locate objects, with categories of objects not being considered and image attributes (such as brightness and scale) unchanged. However, the results of most of this kind of methods has not improved much, but it is also an attempt. In contrast, this paper combines DRL method with the current mainstream object detection methods together to adaptively learn the best image attributes. In addition, there are some other meaningful methods using DRL for basic image processing (e.g., enhancement \cite{park2018distort-and-recover}, recovery \cite{yu2018crafting}). Although these methods can adaptively learn the attributes of images step by step, they are not related to the detection task, but only to meet the visual inspection. Moreover, there are many other works that combine DRL and CV \cite{liang2017deep,yoo2017action,ba2015multiple,he2018merge,huang2017learning}, but all of them are different from the method in this paper. Since human thinking is often a sequential decision-making process, algorithms with deep reinforcement learning methods are closer to human behavior than traditional methods. In short, the image processing method based on deep reinforcement learning is a topic worth studying.

In this paper, an adaptive adjustment strategy of image attribute is learned in the framework of Double DQN \cite{van2016deep} combined with Faster RCNN. Both image quality and detection performance can be improved by applying this strategy. To our best knowledge, the problem being considered in this paper is new, and it is rarely being studied in the literature.

\section{Methodology}
Imaging configuration is an important factor affecting image quality and object detection performance. In particular, brightness and scale are the two most important factors. In addition, the indicators used to evaluate the imaging configuration are different for different tasks, such as visual inspection and object detection. To this end, this paper takes brightness learning and scale learning as examples to study the active imaging configuration learning in the context of object detection task. Below, active object detection (\emph{RL-AOD}) will be formulated, and proposed approach will be elaborated step by step.

\begin{figure}[tb]
  \centering
  \includegraphics[width=8.5cm, height=9cm]{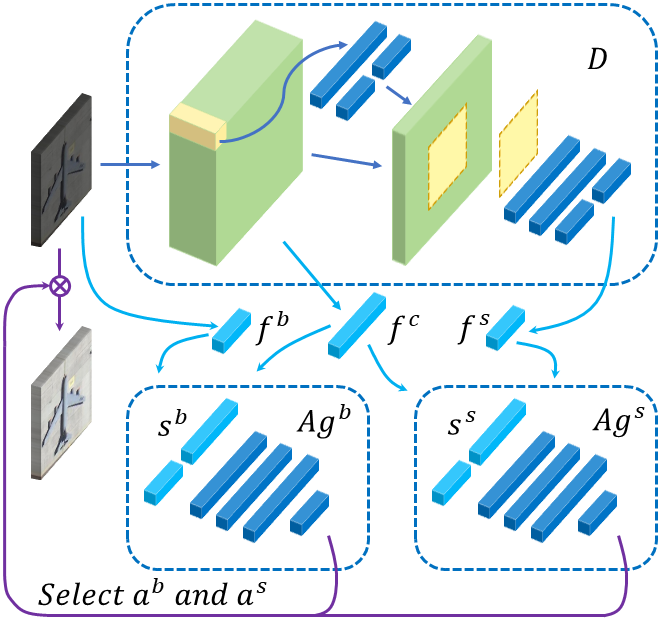}
\caption{Overview of RL-AOD. Firstly, $D$ is used to extract features and detect objects from ${img}(t)$. Then ${Ag}^b$ and ${Ag}^s$ are utilized to select the optimal action $a^b(t)$ and $a^s(t)$ according to the state $s^b(t)$ and $s^s(t)$ respectively. Finally, act is performed on ${img}(t)$ in order to obtain ${img}(t+1)$. }
\label{fig:fig4}
\end{figure}

\subsection{Problem Formulation}
Deep reinforcement learning consists of five key elements, namely environment, agent, state, action and reward. Below, we explain them in the context of image attribute learning.

\textbf{Environment.} The role of environment is to receive a series of actions performed by the agent, to evaluate the quality of these actions, and to feedback reward to the agent. The environment in this article refers to the object detector, abbreviated as $D$. Since the input image size of the one-stage detection method is fixed, it is difficult to adjust the scale. Therefore, this paper uses the Faster RCNN method to construct \emph{RL-AOD} framework. The detector $D$ is trained in advance on a high quality dataset.

\textbf{Agent.} Agent is the core of the entire reinforcement learning system, whose task is to learn a series of state-to-action mappings based on the reward provided by the environment.  The agent in this framework is expected to select the appropriate brightness adjustment actions and scale adjustment actions to transform the image according to the current image feature, and finally adapt to the detector $D$, and improve the overall performance. In brightness adjustment and scale adjustment, two independent agents were trained respectively, named ${Ag}^b$ and ${Ag}^s$.

\begin{algorithm}[tb]
\caption{RL-AOD}
\label{alg:alg1}
\textbf{Input}: Low quality images\\
\textbf{Networks}: Detector $D$, Agent ${Ag}^b$ and ${Ag}^s$\\
\textbf{Parameters}: Feature $f^c$, $f^b$, $f^s$, State $s^b$, $s^s$, Action $a^b$, $a^s$, Action Set $A^b$, $A^s$, Reward $r^b$, $r^s$ \\
\textbf{Output}: High quality images
\begin{algorithmic}[1]
\STATE Pretrain Faster RCNN $D$ on high quality image sets.
\STATE Use $r^b$ calculated by $D$ as a guide to train DQN Agent ${Ag}^b$ on both low and high quality image sets.
\STATE Use $r^s$ calculated by $D$ as a guide to train DQN Agent ${Ag}^s$ on both low and high quality image sets.
\WHILE{ there are still unprocessed images }
\STATE Let step $t=0$.
\STATE Get a low quality image ${img}(0)$
\WHILE{current step $t < T$}
\STATE Extract $f^c(t)$, $f^b(t)$, $f^s(t)$ from ${img}(t)$ using $D$.
\STATE Combine $f^c(t)$, $f^b(t)$ to get $s^b(t)$.
\STATE Combine $f^c(t)$, $f^s(t)$ to get $s^s(t)$.
\STATE Select $a^b(t)$ from $A^b$ based on $s^b(t)$ using ${Ag}^b$.
\STATE Select $a^s(t)$ from $A^s$ based on $s^s(t)$ using ${Ag}^s$.
\STATE Apply $a^b(t)$ and $a^s(t)$ to ${img}(t)$ to get ${img}(t+1)$.
\STATE Step $t$++.
\ENDWHILE
\ENDWHILE
\STATE \textbf{return} High quality image set $\{{img}(T)\}$
\end{algorithmic}
\end{algorithm}

\textbf{State.} State refers to the current status of the agent and contains all the information used to make the action selection. In this paper, the state mainly consists of two parts, one part is the contextual feature of the image, which is used to describe the overall background of the image and denoted as $f^c$. Another part of the feature is used to judge the level of a certain attribute (brightness, scale) of the image, written as $f^b$ and $f^s$, respectively. Therefore, the states corresponding to ${Ag}^b$ and ${Ag}^s$ are $s^b=\{f^c,f^b\}$ and $s^s=\{f^c,f^s\}$, respectively.

\textbf{Action.} Action refers to the best action that the agent picks from the action set $A$. For the brightness-adjusting agent ${Ag}^b$, the action set includes two actions: brightening and darkening. $A^b=\{a^{b}_1,a^{b}_2\}$. Similarly, the scale-adjusting agent ${Ag}^s$, the action set includes two actions: zoom in and zoom out. $A^s=\{a^{s}_1,a^{s}_2\}$. Two agents will select the best action from their respective action sets based on the state.

\textbf{Reward.} Reward is used to evaluate performance of the agent at a certain time step, and it is provided by the environment. In this paper, the reward $r$ is based on the detection performance.
\begin{equation}\label{eq:eq1}
\begin{array}{l}
r(t)=sign(p(t+1)-p(t))\\
\end{array}
\end{equation}
Where $p$ stands for detection performance, and $p=\frac{1}{2}(F+mIoU)$. $mIoU$ is the average $IoU$ of all correct detection boxes, and $F$ is the F-measure of detection boxes with $IoU>0.5$. By experiments, we find that the pure usage of $F$ indicator and $mIoU$ does not work well. $mIoU$ alone is less robust to multiple objects, and $F$ indicator alone often leads to a small reward.

In general, the process of the \emph{RL-AOD} algorithm is reflected in Alg. \ref{alg:alg1} and Fig. \ref{fig:fig4}. After the detector and the two agents are trained separately, $D$ is used to extract features and detect objects of ${img}(t)$. Then features $f^c(t)$, $f^b(t)$ and $f^s(t)$ can be obtained. Next, ${Ag}^b$ and ${Ag}^s$ are utilized to select the optimal action $a^b(t)$ and $a^s(t)$ according to the state $s^b(t)$ and $s^s(t)$ ($s^b=\{f^c,f^b\}$, $s^s=\{f^c,f^s\}$) respectively. Finally, act is performed on ${img}(t)$ in turn, in order to obtain ${img}(t+1)$. The detailed process of feature extraction, state transition, and action design will be described below.

\begin{figure}[tb]
  \centering
  \includegraphics[width=8.5cm]{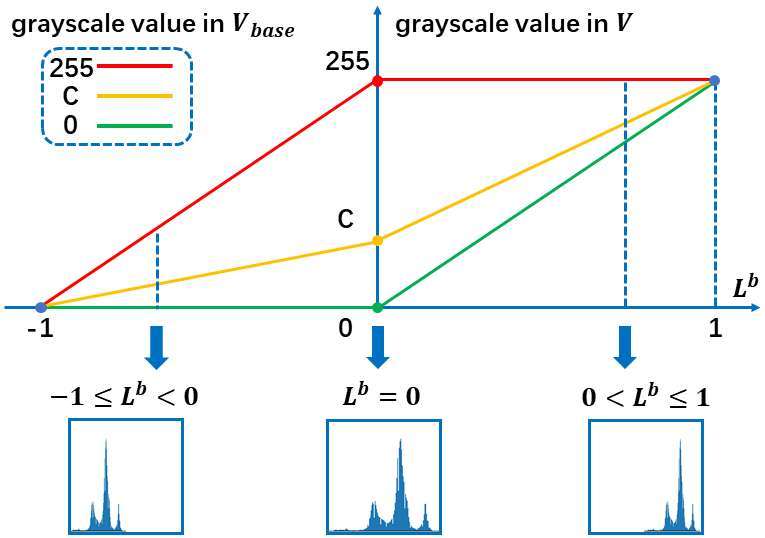}
\caption{Brightness transformation. This figure illustrates how the grayscale value of $V$ is adjusted under different brightness levels $L^b$ when the grayscale value of $V_{base}$ is $0$, $255$ and $C$, respectively. At the same time, the histogram of $V$ under different $L^b$ is also given.}
\label{fig:fig3}
\end{figure}

\subsection{Automatic Brightness Adjustment}
Agent will take the extracted features as the state and take the optimal action to transform the state. Feature extraction part will mainly introduce the acquisition of $f^c$ and $f^b$. State transition part will introduce an indicator $L^b$ for roughly estimating the brightness level of an image. Action design part will propose a set of brightness adjustment actions.

\textbf{Feature extraction.} $f^c$ is the contextual feature extracted by the detector $D$, which is used to describe the overall background of the image. This part of the feature is extracted from the intermediate output of Faster RCNN(feature maps before RoI Pooling layer). If the backbone of the model is based on VGG16, then the feature maps of the intermediate output will have 512 channels, and the averaged feature map over channels will result in a 512-dimensional vector. If the backbone of the model is ResNet50 or ResNet101, this vector will have 1024 dimensions. Experiments show that this dimension is too high and will impact the final performance. Therefore, it is necessary to perform a max-pooling operation with a stride of 2 for the feature, and a feature of 512 dimensions is then obtained. $f^b$ is the histogram feature and is used to judge the level of image's brightness. To compute the histogram, RGB image needs to be converted into HSV color space. Then a histogram is obtained on the component $V$, where the bin width is 4. Since the quantitative level is 256, a 64-dimension histogram can be obtained. Finally, the two parts are concatenated together to obtain $s^b$, a feature vector of 576 dimensions.

\begin{figure}[tb]
\begin{minipage}[b]{.49\linewidth}
  \centering
  \centerline{\includegraphics[width=4.3cm]{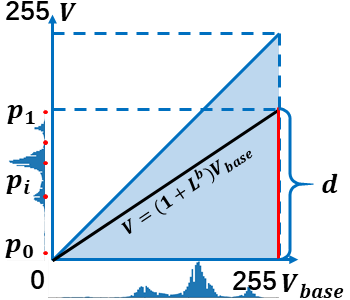}}
  \centerline{(a) }
\end{minipage}
\hfill
\begin{minipage}[b]{.49\linewidth}
  \centering
  \centerline{\includegraphics[width=4.3cm]{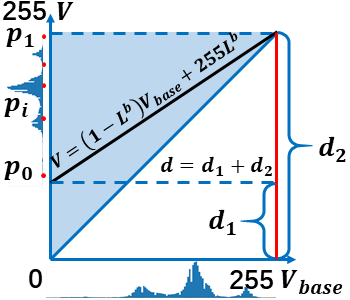}}
  \centerline{(b) }
\end{minipage}
\caption{Geometric meaning of $d$. (a): $-1 \leq L^b < 0$ (Linear mapping between the $V$ component of a dark image and $V_{base}$). (b): $0 \leq L^b \leq 1$ (Linear mapping between the $V$ component of a bright image and $V_{base}$). }
\label{fig:fig2}
\end{figure}

\textbf{State transition.} To describe the brightness level, an indicator $L^b$ is computed on the image brightness component $V$ in HSV color space of RGB image. $L^b$ lies within the range [-1, 1], and changing brightness level $L^b$ means changing the image brightness. The negative number means that the image is dark, and the positive number means that the image is bright. The larger the absolute value, the darker or brighter the image.

To calculate $L^b$, the method in this paper attempts to separate $L^b$ from the brightness component $V$ of the current image. In other words, the brightness component $V$ is separated into two parts. One is the brightness level $L^b$ (a scalar) and the other is called $V_{base}$ (a matrix). Eq. \eqref{eq:eq2} describes the decomposition form.
\begin{equation}\label{eq:eq2}
V(t)=\left\{
\begin{array}{ll}
(1+L^b(t))V_{base}\quad\quad\quad\quad\ {-1 \leq L^b < 0}\\
(1-L^b(t))V_{base}+255L(t)\ \ {0 \leq L^b \leq 1}\\
\end{array} \right.
\end{equation}

In brightness adjustment, only $L^b$ needs to be computed, $V_{base}$ is the constant basis for each image. On this basis, any level of brightness component $V$ can be obtained. The relationship between the grayscale value of $V_{base}$ and the grayscale value of $V$ under different $L^b$ is shown in Fig. \ref{fig:fig3}. So estimating $V_{base}$ is crucial. As shown in Eq. \eqref{eq:eq3}, $L^b(0)$ can be calculated. $V_{base}$ can be got by letting $t=0$ and substituting $L^b(0)$ into Eq. \eqref{eq:eq2}.
\begin{equation}\label{eq:eq3}
L^b(0)=\frac{d}{255}-1,\quad d\approx\frac{\sum_{i}p_i}{6}
\end{equation}
Where $\{p_{0},p_{0.1},p_{0.2},\dots,p_{1}\}$ are $11$ quantiles of the $V$ component, which are indicated by red dots in Fig. \ref{fig:fig2}(a) and (b). $d$ in Eq. \eqref{eq:eq3} is indicated by a solid red line in Fig. \ref{fig:fig2}(a) and (b). The principle of using the quantiles to estimate $d$ can be clearly illustrated in the figure. After pairing the quantiles (such as $\{(p_{0},p_{1}),(p_{0.1},p_{0.9} )\cdots\}$), the sum of each pair is an estimate of $d$, and all estimates are averaged to get a more accurate estimate of $d$, ie, Eq. \eqref{eq:eq3}.

\textbf{Action design.}
The essence of action design is to change the brightness level of the image. The brightness level $L^b(t+1)$ can be updated from $L^b(t)$ by Eq. \eqref{eq:eq4}, and the situation where $L^b$ is outside the range [-1,1] can be avoided. For dark images, the changing scope of $L^b$ is greater in the brightening operation, and for bright images the changing scope of $L^b$ is greater in the darkening operation. In consequence, the agent can choose a good action in the next step even it takes a bad action in this step.
\begin{equation}\label{eq:eq4}
L^b(t+1)=\left\{
\begin{array}{lr}
0.9L^b(t)+0.1\times 1 & {a^b(t)=a^b_1}\\
0.9L^b(t)+0.1\times (-1) & {a^b(t)=a^b_2}\\
\end{array} \right.
\end{equation}

The reason why $V(t+1)$ is not obtained by multiplying $V(t)$ by a coefficient is that in the subsequent brightness adjustment, the truncated grayscale values larger than 255 are difficult to be recovered. In addition, it is easy to enter the situation in which two actions are alternately selected, because if two actions are selected in turn, state will return to origin.

\subsection{Automatic Scale Adjustment}
Agent will take the extracted features as the state and take the optimal action to transform the state. Feature extraction part will mainly introduce the acquisition of $f^s$. State transition part will introduce an indicator $L^s$ for roughly estimating the scale level of an image. Action design part will propose a set of scale adjustment actions.

\textbf{Feature extraction.}
$f^s$ is the statistical histogram of objects' area. Firstly, the detector $D$ is used to detect the image, and a series of bounding boxes are obtained. On the area of these bounding boxes, a histogram is obtained. Because it is the area that is being counted, the design of the bin width will show a tendency to widen, and the speed of the increase will be square. The specific form of bin is [$0$, $9^2$, $10^2$ $\cdots$ $24^2$, $27^2$ $\cdots$ $75^2$, $80^2$ $\cdots$ $175^2$, $182^2$ $\cdots$ $245^2$, +$\infty$]. After manual design, this feature $f^s$ will be 64-dimensional. Since not every image has a lot of objects, and some images don't even contain any objects, the extracted histogram feature is a very sparse vector. This is very unfavorable for subsequent training. This sparse feature can be convolved by a Gaussian kernel to make it less sparse. Similarly, $f^c$ and $f^s$ are concatenated together to obtain $s^s$, a 576-dimension vector.

\textbf{State transition.} Similarly, the scale level $L^s$ is defined for adjusting the size of images. $L^s$ lies within the range [-1, 1], and changing scale level $L^s$ means changing the size of images. It is worth noting that the resolution of the image is not absolutely related to the scale level $L^s$. The scale level $L^s$ can only describe the average area of all objects in an image. The negative number means that the average area of all objects in this image is small, and the positive number means the average area is large. The larger the absolute value, the smaller or larger the average area. In step $t$, bilinear interpolation can be used to resize the image $img_t^s$ to $\theta^{L^s(t)}$ times of the size to get a new image $img_{t+1}^s$, as expressed by Eq. \eqref{eq:eq5}. After $T$ steps, the scale of the new image is $\theta^{\sum_{t}^{T}L^s(t)}$ times that of the original image.
\begin{equation}\label{eq:eq5}
img^s(t+1)=Resize(img^s(t), \theta^{L^s(t)})
\end{equation}

After defining the scale level $L^s$, it is now necessary to determine the $L^s$ corresponding to the original image, that is, estimating $L^s(0)$. Eq. \eqref{eq:eq6} shows the estimation process.
\begin{equation}\label{eq:eq6}
L^s(0)=\frac{1}{2}{log}_{\theta}(\frac{\alpha}{\alpha_0})
\end{equation}
Where $\alpha$ refers to the average area of objects. Both $\alpha_0$ and $\theta$ are auxiliary parameters. $\alpha_0$ represents a priori average area with a value of $96^2$. This is the threshold for area of medium-size objects and large-size objects in the COCO dataset evaluation criteria. $\theta$ is set to $8$. With this setting, images with the average area of objects between $16^2$ and $768^2$ correspond to $L^s$ in [-1, 1], and the average area outside the range will give $L^s$ a value of 1 or -1.

\textbf{Action design.}
$L^s$ is adjusted in the similar way with Eq. \eqref{eq:eq4}. Eq. \eqref{eq:eq7} is the adjustment method of $L^s$.
\begin{equation}\label{eq:eq7}
L^s(t+1)=\left\{
\begin{array}{lr}
0.95L^s(t)+0.05\times 1 \quad \ \ \ {a^s(t)=a^s_1}\\
0.95L^s(t)+0.05\times (-1) \ {a^s(t)=a^s_2}\\
\end{array} \right.
\end{equation}
In this way, $L^s$ can be avoided going beyond the range [-1,1]. For images have large average area of objects, the changing scope of $L^s$ will be greater in the zoom out operation, and for images have small average area of objects, the changing scope of $L^s$ is greater in the zoom in operation. In consequence, even if the agent takes a bad action in this step, it can still choose a good action in the next step.

\section{Experiments}
\subsection{Datasets and Setting}
In order to verify the effectiveness of the method, this paper carried out experiments on a remote sensing image dataset. The details of the dataset and the settings of parameters in the network will be introduced next.

\textbf{Dataset.} Our interests focus mainly on aircraft, and we collect many very high resolution remote sensing images and build a dataset for aircraft detection. The dataset is consisted of $13,078$ training images and $5,606$ test images. These images are high quality images obtained under normal conditions. In fact, due to the complexity of the environment, it is impossible to always get high quality images through imaging devices in the real world. For example, if the environment is too dark, the quality of the image obtained must not be high. In this paper, the degradation of the test set based on brightness and scale is performed by simulation. As shown in Fig. \ref{fig:fig1}, each test image can be subjected to four degradation operations, and thereby five images can be obtained. In this way, $28,030$ test images can be obtained totally. In the following, all models are trained in the degraded dataset.

\textbf{Settings.} In this paper, the detector refers to Faster RCNN model. This model is trained in the training set containing $13,078$ images mentioned above. Whether the model backbone is VGG16, ResNet50 or ResNet101, the scales of the anchor is set to $(4, 8, 16, 32)$, and the number of iterations is set to $110000$. The rest of the settings are set according to the model default settings. The structures of the agents corresponding to brightness and scale adjustment are exactly the same, both using Double DQN model. Double DQN is a variant of the DQN algorithm, which is used to eliminate the overestimation of Q-values and is more stable. The agent network is a six-layer fully connected neural network. The neurons in each layer are $512, 512, 512, 512, 512, 2$. When training agent networks, $13,078$ training images are randomly degraded (changing brightness and scale). The degraded images account for 80\% of the total training image set. Adam optimizer is used for agent network learning with a basic learning rate of $0.001$. The brightness-adjusted agent network will train $120,000$ iterations, and the scale-adjusted agent network trains $40,000$ iterations. During the training process, the action selection adopts the $\epsilon$-greedy method. This method will randomly select actions with a small probability, and select the optimal action in the rest.

\begin{table}[tb]
\small
\centering
\caption{Performance comparison of different methods.}
\setlength{\tabcolsep}{0.9mm}{
\begin{tabular}{lcccccc}
  \toprule
  Method+Backbone & $AP$ & $AP^{50}$ & $AP^{75}$ & $AP^S$ & $AP^M$ & $AP^L$ \\
  \midrule
  DPMv5 (benchmark) & - & 0.338 & - & - & - & - \\
  \midrule
  RetinaNet+VGG16 & 0.376 & 0.585 & 0.431 & 0.249 & 0.492 & 0.526 \\
  RetinaNet+ResNet50 & 0.446 & 0.674 & 0.515 & 0.297 & 0.591 & 0.588 \\
  RetinaNet+ResNet101 & 0.503 & 0.732 & 0.596 & 0.338 & 0.654 & 0.681 \\
  SSD321+ResNet101 & 0.417  & 0.661  & 0.475  & 0.200  & 0.594  & 0.703 \\
  DSSD321+ResNet101 & 0.426  & 0.666  & 0.485  & 0.196  & 0.610  & 0.739 \\
  YOLOv2+DarkNet19 & 0.407 & 0.632 & 0.472 & 0.202 & 0.573 & 0.701 \\
  YOLOv3+DarkNet53 & 0.491  & 0.808  & 0.553  & \textbf{0.441}  & 0.574  & 0.401 \\
  \midrule
  R-FCN+ResNet50 & 0.422  & 0.705  & 0.461  & 0.223  & 0.576  & 0.690 \\
  R-FCN+ResNet101 & 0.427  & 0.713  & 0.464  & 0.225  & 0.582  & 0.694 \\
  Faster RCNN+VGG16 & 0.441  & 0.750  & 0.469  & 0.273  & 0.587  & 0.652 \\
  Faster RCNN+Res50 & 0.455  & 0.768  & 0.482  & 0.273  & 0.601  & 0.700 \\
  Faster RCNN+Res101 & 0.479  & 0.784  & 0.525  & 0.300  & 0.626  & 0.703 \\
  \midrule
  RL-AOD+VGG16 & 0.530  & \textbf{0.822}  & 0.608  & 0.355  & \textbf{0.674}  & \textbf{0.751} \\
  RL-AOD+ResNet50 & 0.519  & 0.815  & 0.590  & 0.346  & 0.661  & 0.734 \\
  RL-AOD+ResNet101 & \textbf{0.531}  & \textbf{0.822}  & \textbf{0.612}  & 0.361  & 0.664  & 0.750 \\
\bottomrule
\end{tabular}}
\label{tab:tab1}
\end{table}

\begin{table}[tb]
\normalsize
\centering
\caption{Performance comparison of different parameter settings of \emph{RL-AOD}. FR refers to the original Faster RCNN method. $B$ refers to the brightness adjustment. $S$ refers to the scale adjustment. $2$ and $4$ represent the maximum step $T$ ($T$ is defined in Alg. \ref{alg:alg1}). $\ast$ represents the result of testing on the undamaged normal dataset.}
\setlength{\tabcolsep}{0.9mm}{
\begin{tabular}{lcccccc}
  \toprule
  Method+Backbone & $AP$ & $AP^{50}$ & $AP^{75}$ & $AP^S$ & $AP^M$ & $AP^L$ \\
  \midrule
  FR+VGG16 & 0.441  & 0.750  & 0.469  & 0.273  & 0.587  & 0.652 \\
  B2+VGG16 & 0.498  & 0.821  & 0.542  & 0.312  & 0.650  & 0.741 \\
  BS2+VGG16 & 0.515  & 0.811  & 0.585  & 0.339  & 0.662  & 0.736 \\
  B4+VGG16 & 0.503  & 0.823  & 0.554  & 0.314  & 0.655  & 0.754 \\
  BS4+VGG16 & 0.530  & 0.822  & 0.608  & 0.355  & 0.674  & 0.751 \\
  \midrule
  FR+Res50 & 0.455  & 0.768  & 0.482  & 0.273  & 0.601  & 0.700 \\
  B2+Res50 & 0.499  & 0.825  & 0.538  & 0.308  & 0.646  & 0.744 \\
  BS2+Res50 & 0.509  & 0.805  & 0.569  & 0.333  & 0.650  & 0.730 \\
  B4+Res50 & 0.508  & 0.831  & 0.552  & 0.318  & 0.656  & 0.753 \\
  BS4+Res50 & 0.519  & 0.815  & 0.590  & 0.346  & 0.661  & 0.734 \\
  \midrule
  FR+Res101 & 0.479  & 0.784  & 0.525  & 0.300  & 0.626  & 0.703 \\
  B2+Res101 & 0.510  & 0.824  & 0.558  & 0.322  & 0.657  & 0.755 \\
  BS2+Res101 & 0.524  & 0.813  & 0.602  & 0.352  & 0.661  & 0.745 \\
  B4+Res101 & 0.514  & 0.833  & 0.565  & 0.324  & 0.660  & 0.763 \\
  BS4+Res101 & 0.531  & 0.822  & 0.612  & 0.361  & 0.664  & 0.750 \\
  \midrule
  FR$\ast$+Res101 & 0.574  & 0.875  & 0.649  & 0.401  & 0.711  & 0.785 \\
  BS4$\ast$+Res101 & 0.575  & 0.873  & 0.651  & 0.402  & 0.711  & 0.786 \\
\bottomrule
\end{tabular}}
\label{tab:tab2}
\end{table}

\begin{figure*}[thb]
\begin{minipage}[t]{0.16\linewidth}
  \centering
  \includegraphics[width=2.6cm]{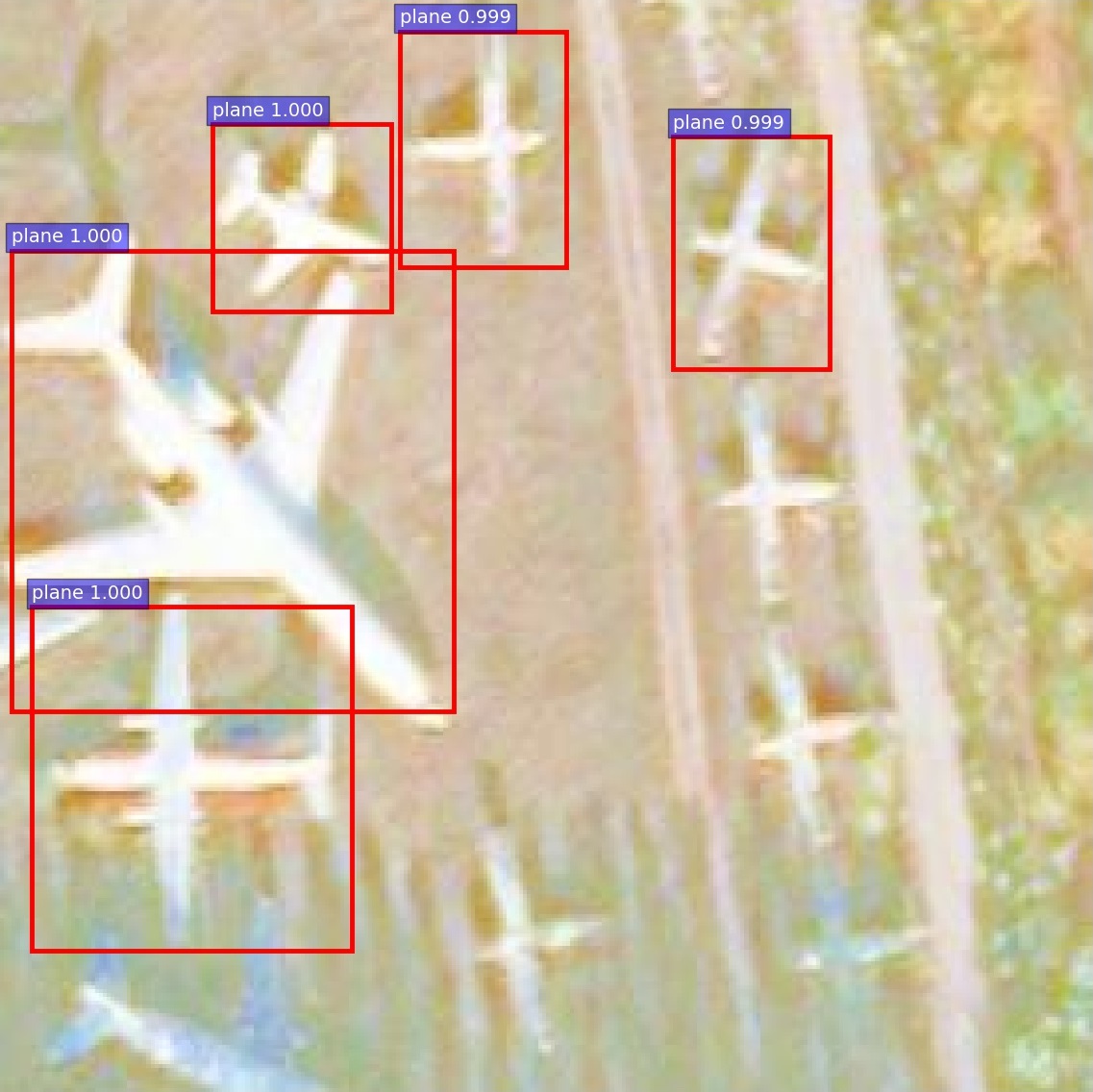}
  \centerline{(a)}
\end{minipage}
\begin{minipage}[t]{0.16\linewidth}
  \centering
  \includegraphics[width=2.6cm]{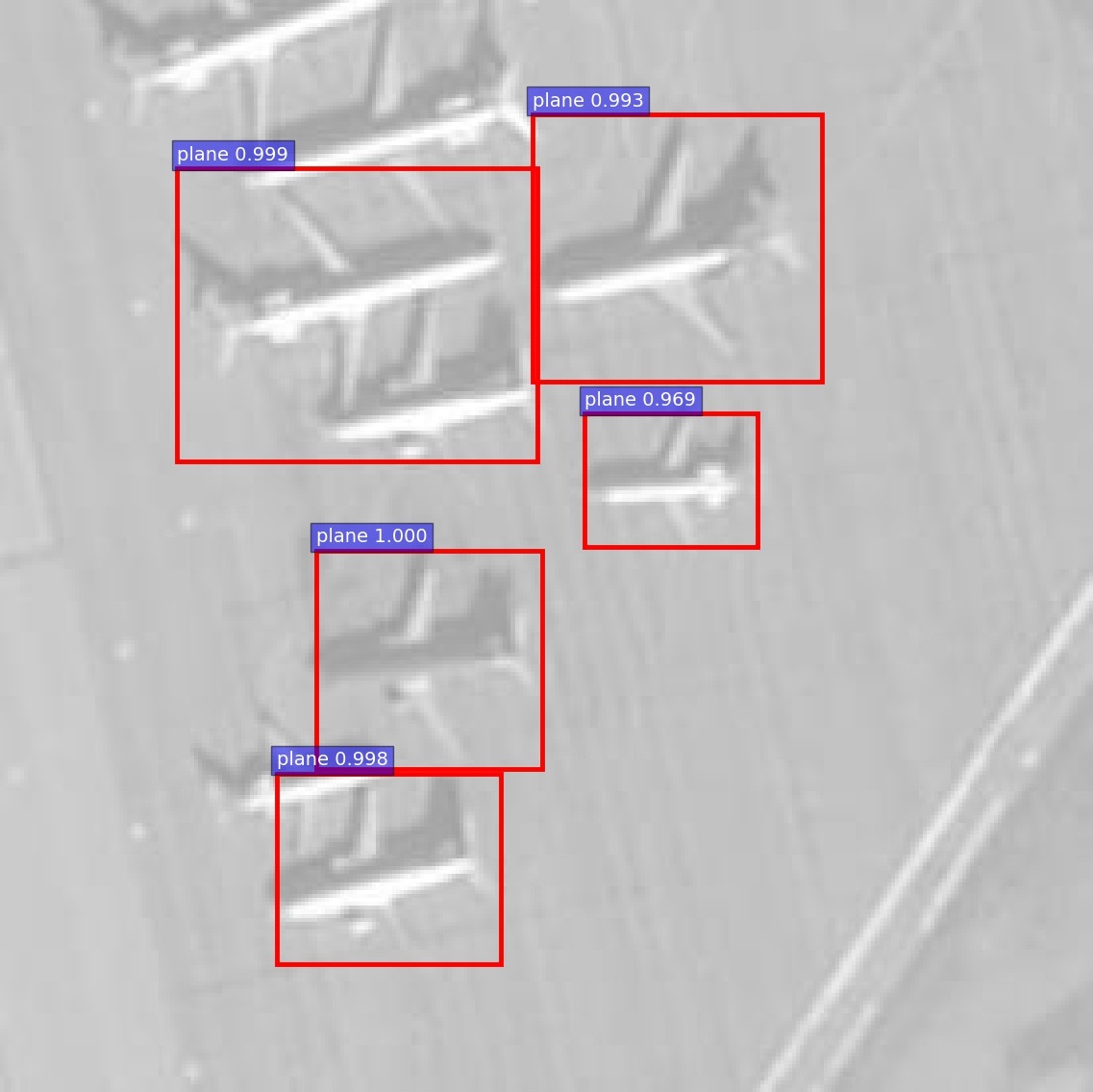}
  \centerline{(b)}
\end{minipage}
\begin{minipage}[t]{0.16\linewidth}
  \centering
  \includegraphics[width=2.6cm]{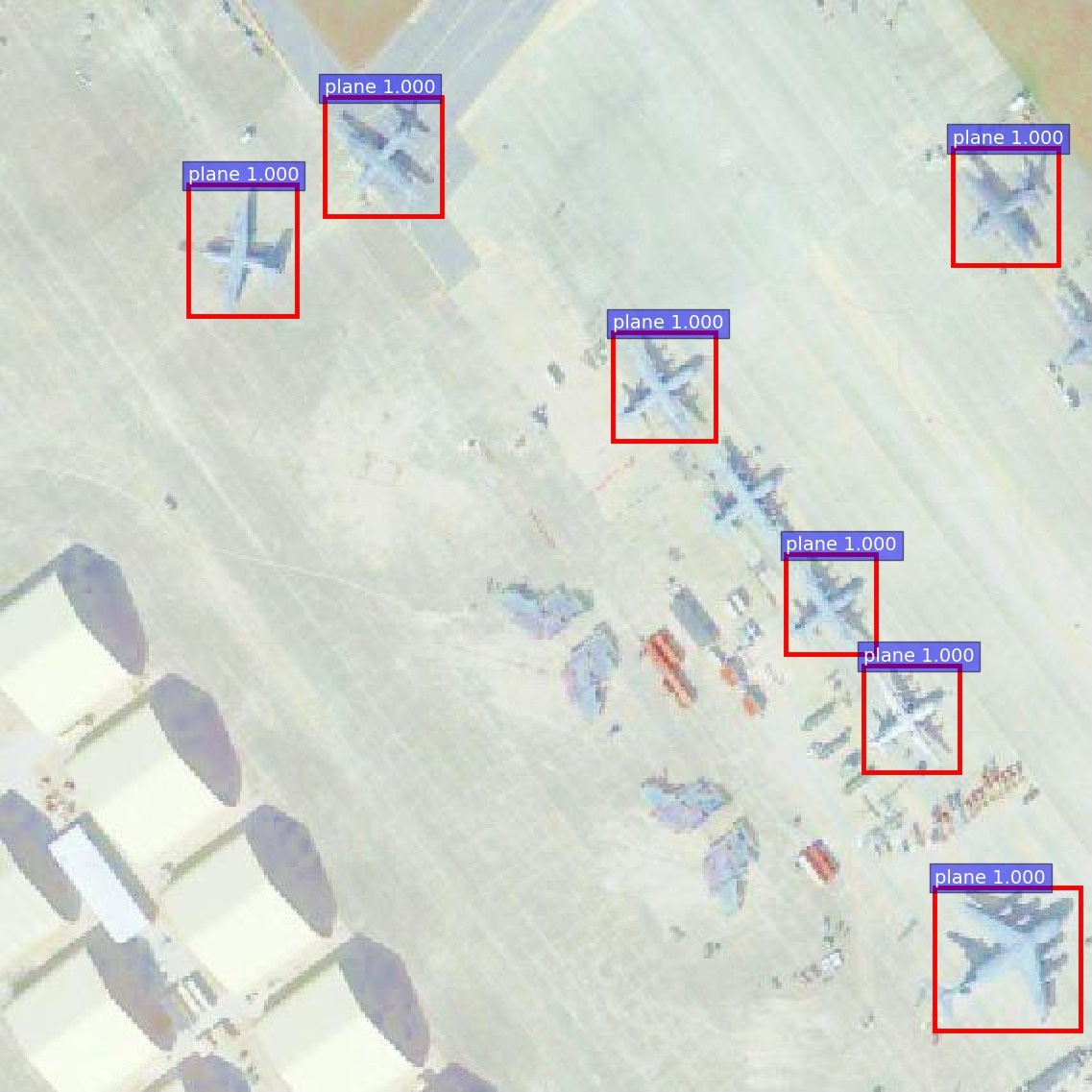}
  \centerline{(c)}
\end{minipage}
\begin{minipage}[t]{0.16\linewidth}
  \centering
  \includegraphics[width=2.6cm]{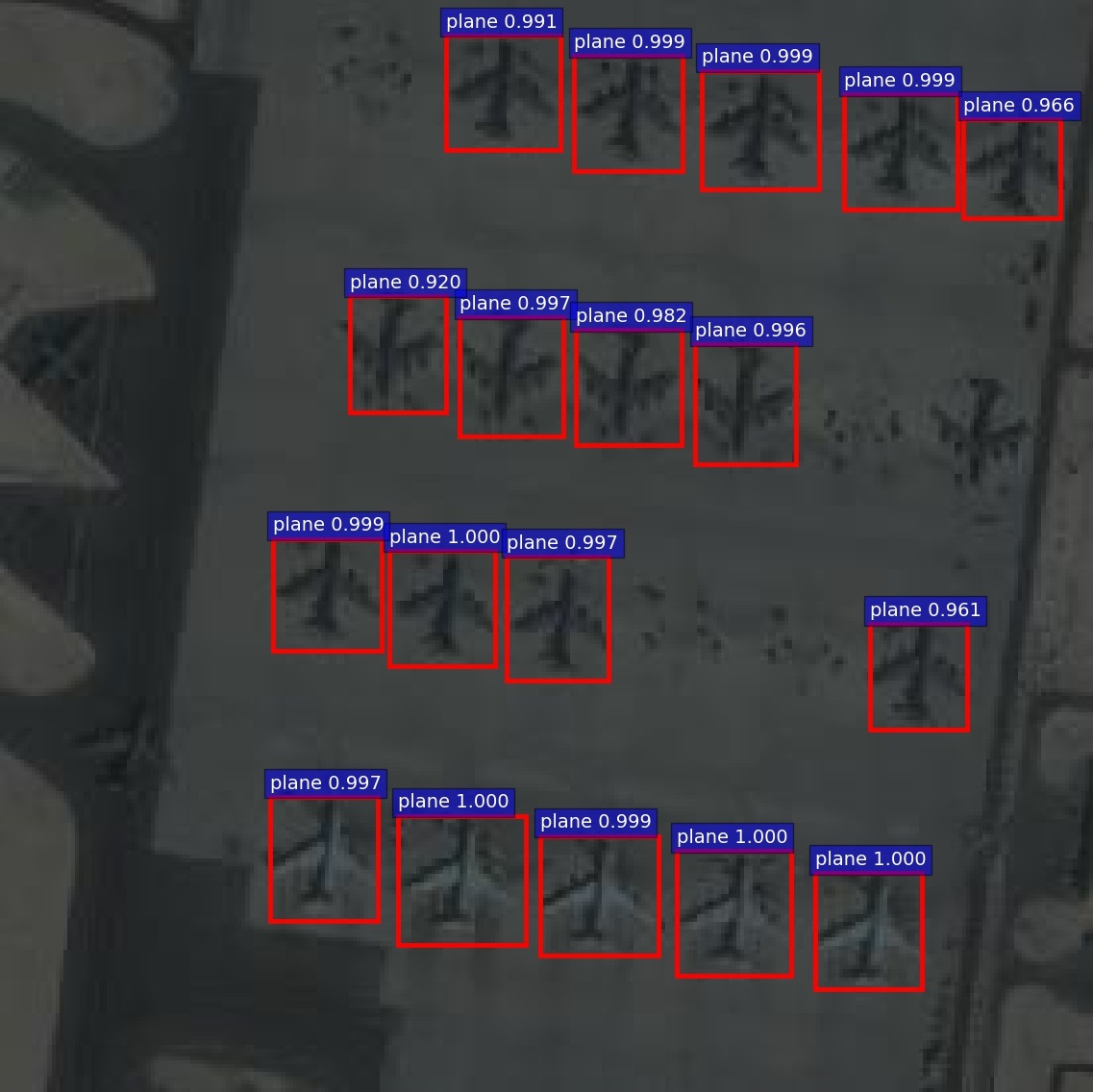}
  \centerline{(d)}
\end{minipage}
\begin{minipage}[t]{0.16\linewidth}
  \centering
  \includegraphics[width=2.6cm]{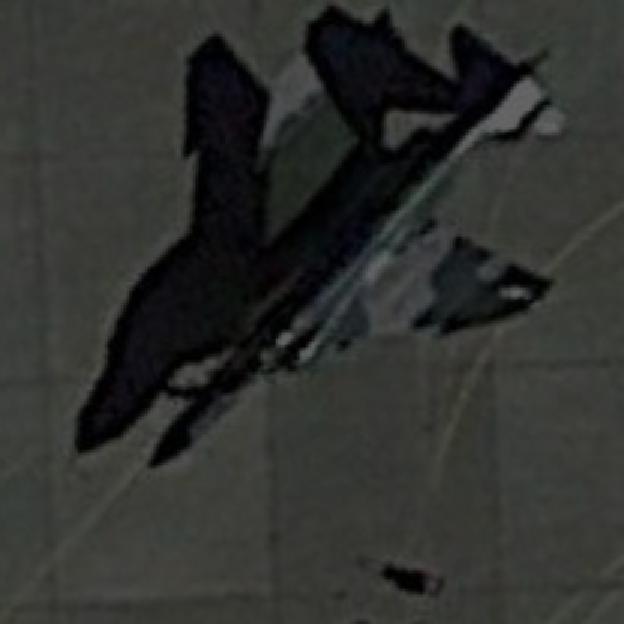}
  \centerline{(e)}
\end{minipage}
\begin{minipage}[t]{0.16\linewidth}
  \centering
  \includegraphics[width=2.6cm]{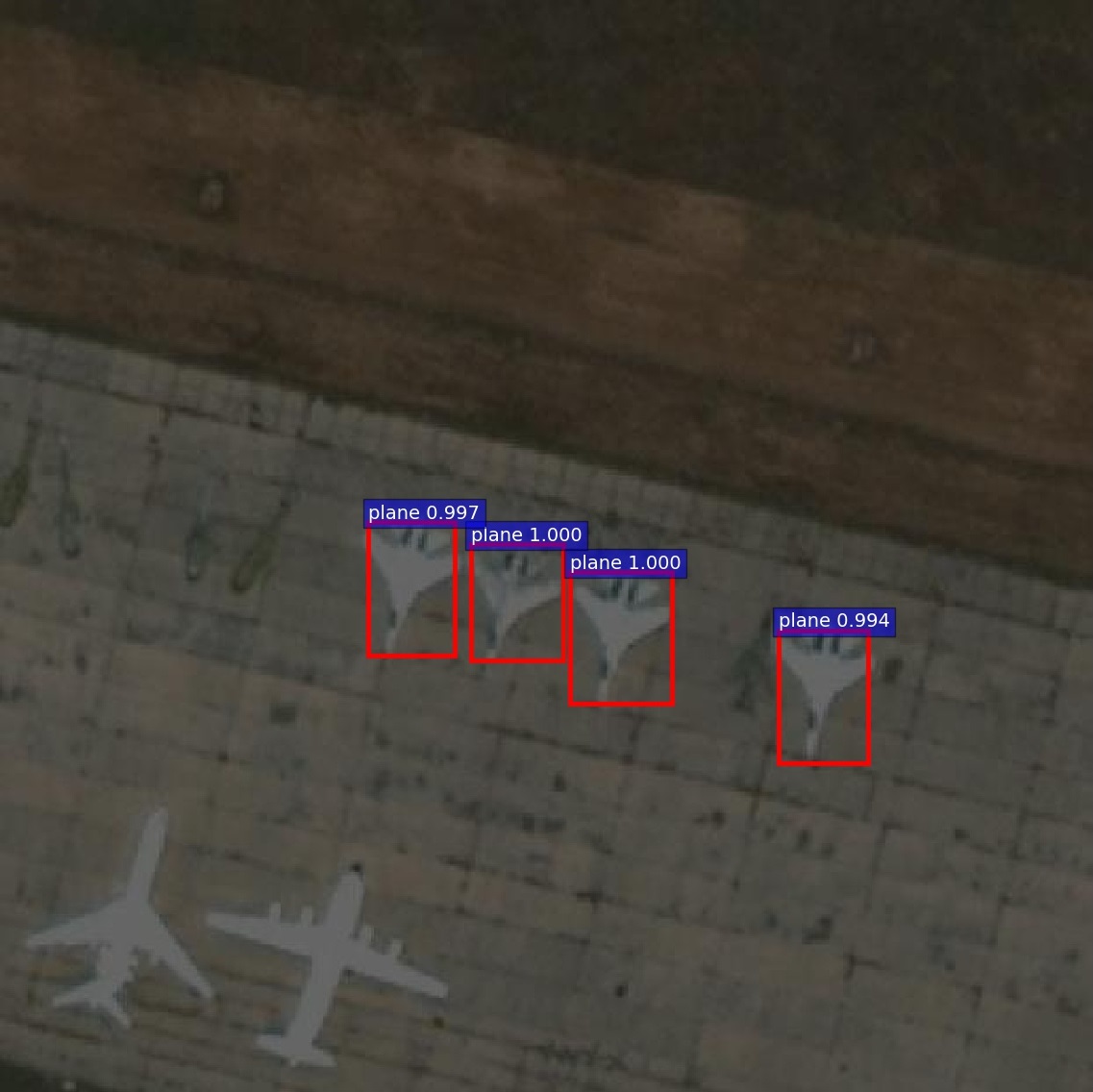}
  \centerline{(f)}
\end{minipage}
\\ \protect\\ \\
\begin{minipage}[t]{0.132\linewidth}
  \centering
  \includegraphics[width=2.1cm]{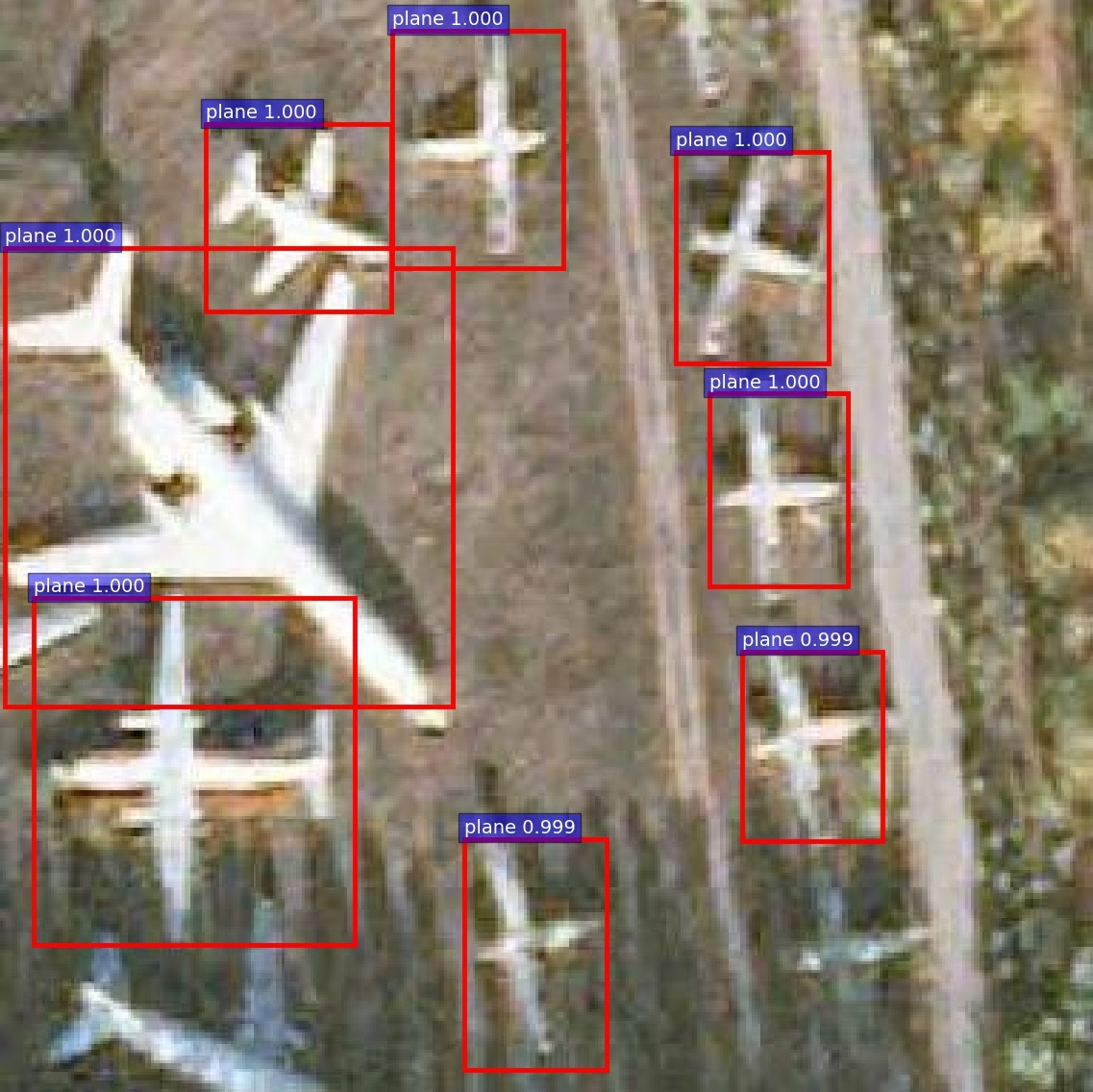}
  \centerline{(g)}
\end{minipage}
\begin{minipage}[t]{0.132\linewidth}
  \centering
  \includegraphics[width=2.1cm]{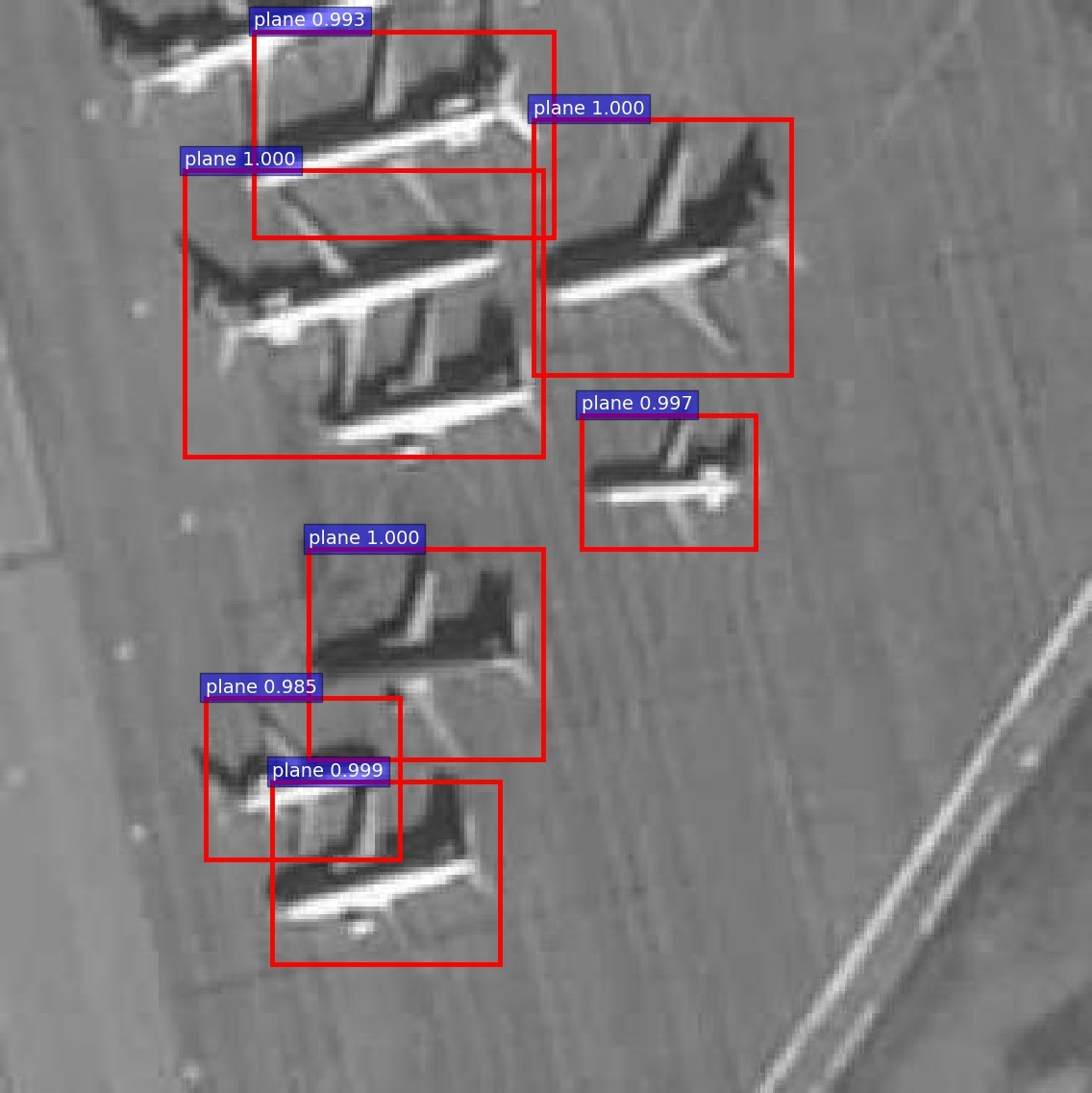}
  \centerline{(h)}
\end{minipage}
\begin{minipage}[t]{0.206\linewidth}
  \centering
  \includegraphics[width=3.4cm]{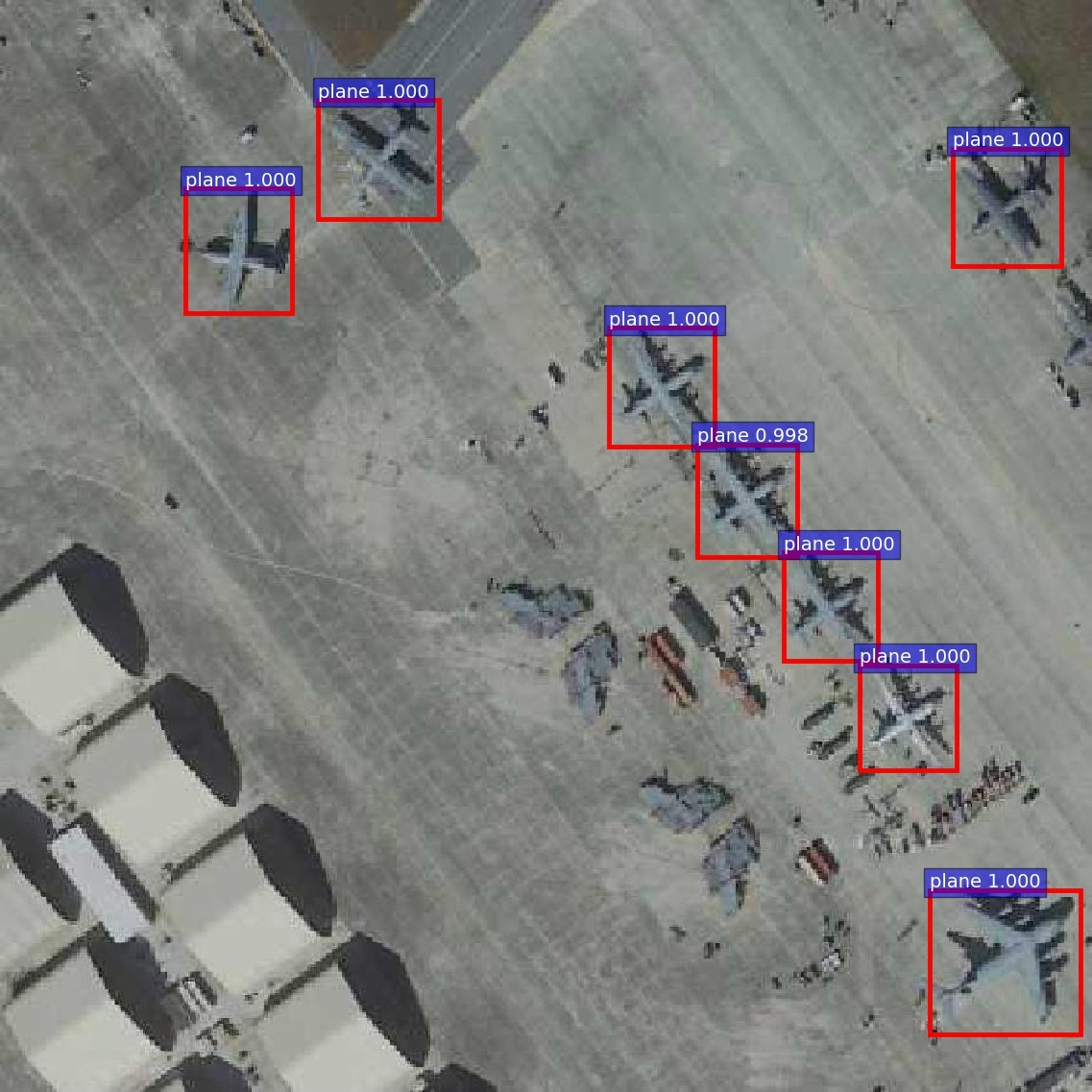}
  \centerline{(i)}
\end{minipage}
\begin{minipage}[t]{0.206\linewidth}
  \centering
  \includegraphics[width=3.4cm]{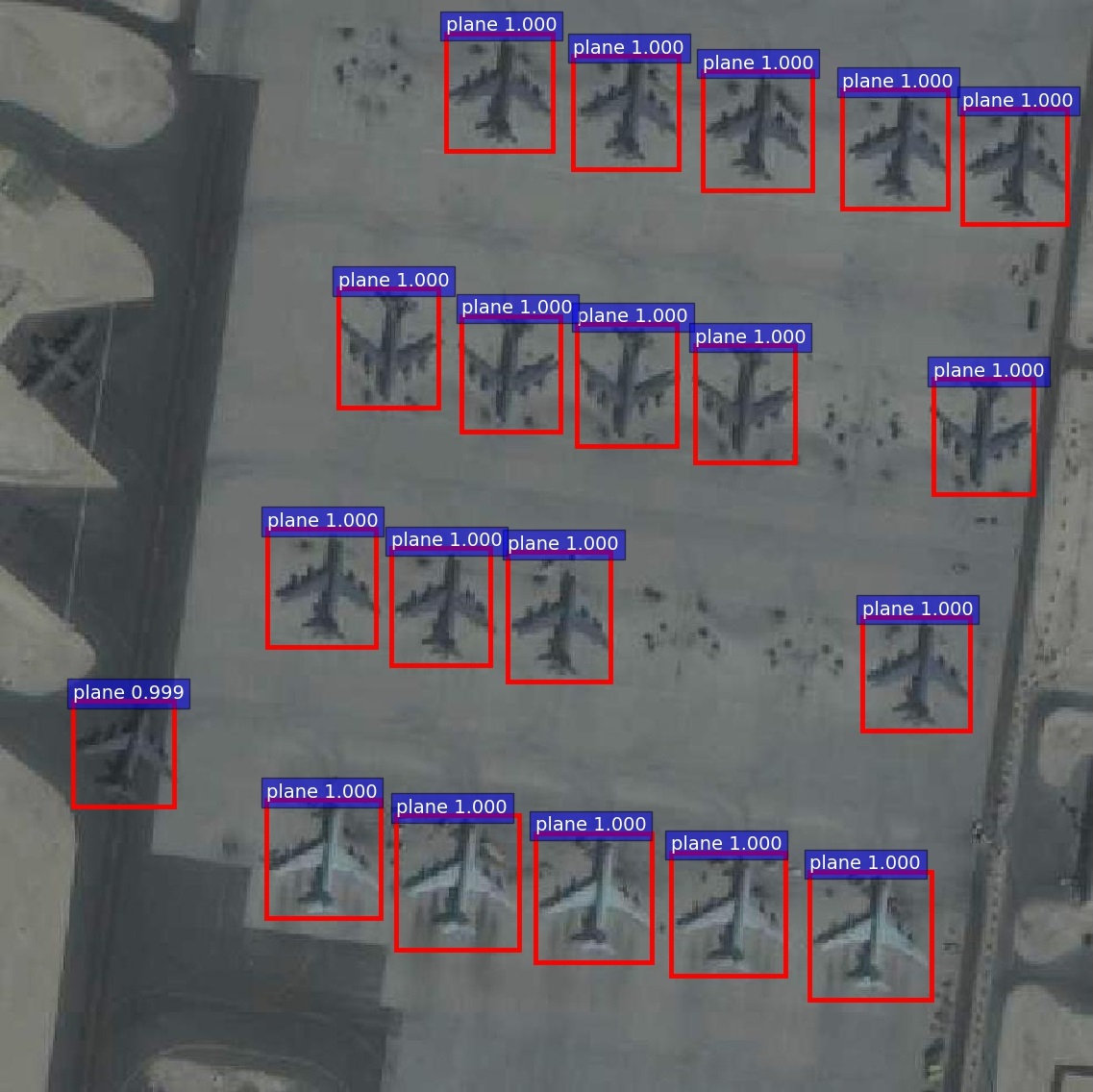}
  \centerline{(j)}
\end{minipage}
\begin{minipage}[t]{0.102\linewidth}
  \centering
  \includegraphics[width=1.6cm]{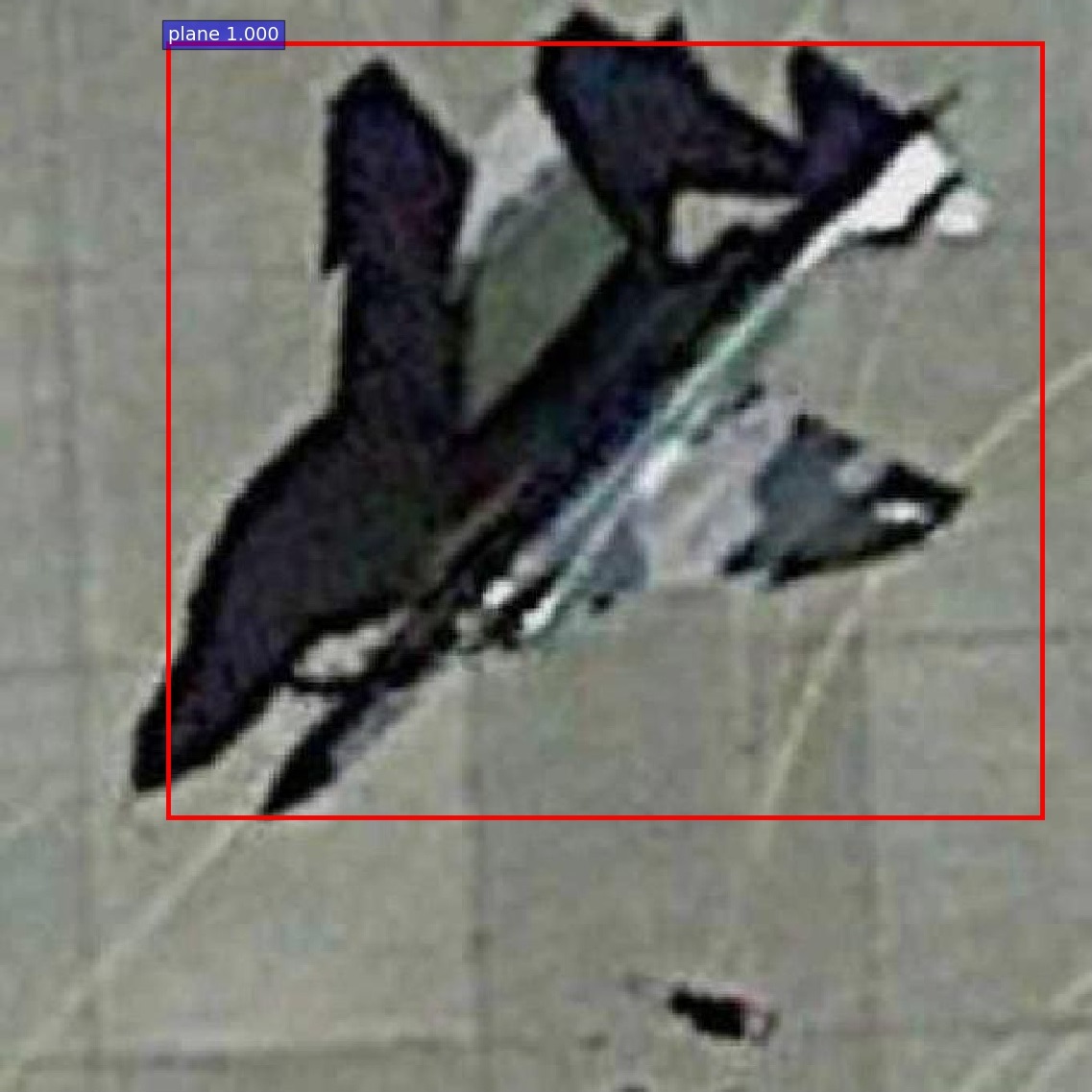}
  \centerline{(k)}
\end{minipage}
\begin{minipage}[t]{0.183\linewidth}
  \centering
  \includegraphics[width=3cm]{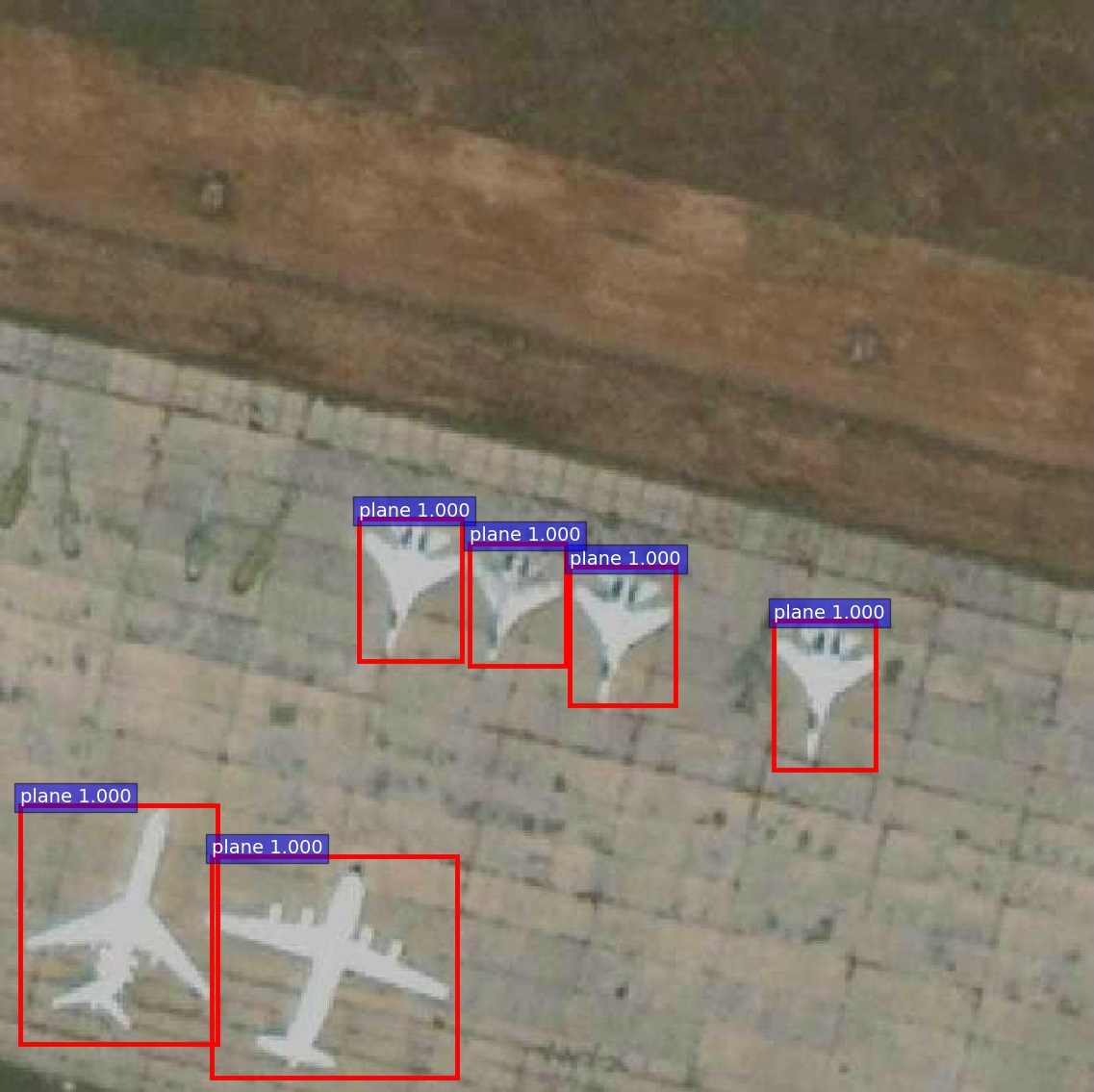}
  \centerline{(l)}
\end{minipage}
\caption{Results comparison. The images (a, b, c, d, e, f) in the first row are the detection results on the degraded images with respect to brightness and scale. The images (g, h, i, j, k, l) in the second row are the results after adaptive learning by \emph{RL-AOD}. (a, b, c) are to simulate the situation of over-exposure, and (d, e, f) are to simulate the situation of under-exposure. \emph{RL-AOD} can find the best image attributes (brightness and scale) through the sequence decision method, so that the detection performance can be improved.}
\label{fig:fig5}
\end{figure*}

\subsection{Results and Discussions}
In this section, \emph{RL-AOD} will be compared to different state-of-the-art methods, and \emph{RL-AOD} with different parameter settings will be also compared. The following will be combined with Fig. \ref{fig:fig5}, Tab. \ref{tab:tab1} and Tab. \ref{tab:tab2} for analysis.

\textbf{Comparison of different methods.} The performances of the different methods are listed in Tab. \ref{tab:tab1}. DPM is a classic non-deep-learning object detection method, selected as a benchmark over feature-detector-descriptor. It can be observed that \emph{RL-AOD}+ResNet101 method has the largest $AP$ of $0.531$. \emph{RL-AOD} based on VGG16 and ResNet50 are also superior to other classical methods, such as Faster RCNN, YOLO, SSD. In order to ensure the fairness of the comparison, all methods, including the mainstream method and the method in this paper, are trained and tested on the damaged data set. Therefore, \emph{RL-AOD} method can achieve better performances in a dataset where the imaging configuration does not match the detector. Since the detector used by \emph{RL-AOD} is Faster RCNN, the two methods can be further compared. With the same backbone, \emph{RL-AOD} outperforms Faster RCNN by $25.7\%$, $10.3\%$ and $8.9\%$ with respect to the averaged indicators $AP^S$, $AP^M$ and $AP^L$. It can be indicated that \emph{RL-AOD} is the most promising for detecting small objects. Faster RCNN performs multiple poolings in the feature extraction process, and rounding in RoI pooling layer leads to a precision loss. In consequence, Faster RCNN is limited in detecting small objects. In contrast, \emph{RL-AOD} compensates for this defect to some extent through adaptive scale adjustment. Although YOLOv3+DarkNet53 is optimal in detecting small objects, it has the lowest detection accuracy of $0.401$ for large objects. Among all methods, \emph{RL-AOD}+ResNet101 is sub-optimal in detecting small objects. Some results of \emph{RL-AOD} are shown in Fig. \ref{fig:fig5}. It can be found that after adaptive attribute learning, many missed objects can be detected. In Fig. \ref{fig:fig5}, images (a, b, c) are to simulate the situation of over-exposure, and images (d, e, f) are to simulate the situation of under-exposure. From the results in Fig. \ref{fig:fig5}, it can be seen that both over-exposure and under-exposure will cause missed alarm, thereby reducing performance. Images (g, h, i, j, k, l) show the results of adaptive learning by \emph{RL-AOD} in this paper. The number of missed alarms has been significantly reduced, and the performance has been improved. In summary, based on the traditional method, \emph{RL-AOD} gradually adjusts the image attributes (brightness and scale) with the help of deep reinforcement learning, so that the damaged image with poor detection effect is more suitable for detection. This is a very meaningful work for remote sensing images.

\textbf{Comparison of different parameter settings.} To understand how \emph{RL-AOD} works, the testing images are evaluated by setting different parameters. The performances are listed in Tab. \ref{tab:tab2}. Firstly, with the same backbone, the performance improvement caused by the brightness adjustment is greater than that brought by the scale adjustment. For example, the difference between AP of B4+Res101 and FR+Res101 ($0.035$) is greater than the difference between AP of BS4+Res101 and B4+Res101 ($0.017$). In brightness and scale adjustment, averaged AP improvements are $0.050$ and $0.018$ respectively, at the maximum step $T$ of $4$. Secondly, as for the performance of the maximum step size of $4$, all indicators are better than the result in the maximum step $T$ of $2$. This shows that \emph{RL-AOD} is effective in adjusting the attributes of images step by step, making them more adaptable to the detector. Thirdly, the sequence decision operation does not reduce detection accuracy of normal images that are not damaged. It can be seen from Tab. \ref{tab:tab2} that the AP values of FR$\ast$+Res101 and BS4$\ast$+Res101 are very close, 0.574 and 0.575 respectively. Fourthly, compared to FR+Res101 and FR$\ast$+Res101, when images are damaged and not suitable for detection, the performance of faster rcnn will be greatly reduced, and the AP will drop by 9.5 points. This also confirms the fact that during the orbit imaging process, if the image acquisition process does not take into account the specific requirements of object detection and other tasks, and the evaluation is not carried out, the detection effect may not be optimal. Finally, it is worth noting that although the scale adjustment is able to improve the detection performance of small and medium-size objects, it will reduce the detection accuracy of some large-size objects. The reason is that in remote sensing image dataset, small and medium-size objects occupy the majority. Generally, it is easier to improve the performance by zoom in than zoom out. Therefore, this will cause the imbalance between large and small-size objects. When training the agent, the agent will be more inclined to enlarge the image. In general, however, the advantages of \emph{RL-AOD} outweigh the disadvantages, as AP is still improved. The introduction of serialized decision-making methods, such as deep reinforcement learning, makes it possible to perform detection while adjusting image attributes.

\section{Conclusion}
This paper proposes an active object detection method \emph{RL-AOD}, which uses deep reinforcement learning to help the object detection module actively adjust image attributes (such as brightness and scale). Traditional object detection methods are limited due to the passive nature, but our active method in this paper can adapt to various situations (such as insufficient brightness, etc.). Experiments demonstrate the necessity of adaptive brightness and scale adjustment, and the effectiveness of \emph{RL-AOD} . Future work will focus on DDPG that produces continuous actions. At the same time, it will also pay more attention to real-time and improve model speed.

\section*{Acknowledgments}
This research was supported by the Major Project for New Generation of AI under Grant No. $2018AAA0100400$, and the National Natural Science Foundation of China under Grants $62071466$ and $91646207$.

\newpage
\bibliographystyle{IEEEtran}
\bibliography{icpr2020}

\end{document}